\documentclass[10pt,twocolumn,letterpaper]{article}

\usepackage{cvpr}
\usepackage{times}
\usepackage{epsfig}
\usepackage{graphicx}
\usepackage{amsmath}
\usepackage{amssymb}

\usepackage{amsthm}
\usepackage{lineno}
\usepackage{multirow}
\usepackage{array}
\usepackage{bm}
\usepackage{paralist}

\usepackage{subfig}
\usepackage{sidecap}
\usepackage{float}
\usepackage{epstopdf}
\usepackage{caption}
\usepackage{mathptmx} 	    
\usepackage{pslatex}
\usepackage{times}

\definecolor{gray}{rgb}{0.75,0.75,0.75}

\newcommand{\comment}[1]{}

\newcommand{\suchthat}{% 
\mathrel{\ooalign{$\ni$\cr\kern-1pt$-$\kern-6.5pt$-$}}} 

\newcommand{\ednote}[1]{{\color{red}#1}}

\DeclareMathOperator*{\argmax}{argmax}

\cvprfinalcopy % *** Uncomment this line for the final submission

\def\cvprPaperID{1409} % *** Enter the CVPR Paper ID here

\ifcvprfinal\fi

\begin{document}

\title{Multi-Region Probabilistic Dice Similarity Coefficient using the Aitchison Distance and Bipartite Graph Matching}

\author{Shawn Andrews\\
Simon Fraser University\\
{\tt\small sda56@sfu.ca}
% For a paper whose authors are all at the same institution,
% omit the following lines up until the closing ``}''.
% Additional authors and addresses can be added with ``\and'',
% just like the second author.
% To save space, use either the email address or home page, not both
\and
Ghassan Hamarneh\\
Simon Fraser University\\
{\tt\small hamarneh@sfu.ca}
}

\maketitle

%\section{}
%\subsection{}

%\section{abstract}
\begin{abstract}
Validation of image segmentation methods is of critical importance. Probabilistic image segmentation is increasingly popular as it captures uncertainty in the results. Image segmentation methods that support multi-region (as opposed to binary) delineation are more favourable as they capture interactions between the different objects in the image. The Dice similarity coefficient (DSC) has been a popular metric for evaluating the accuracy of automated or semi-automated segmentation methods by comparing their results to the ground truth. In this work, we develop an extension of the DSC to multi-region probabilistic segmentations (with unordered labels). We use bipartite graph matching to establish label correspondences and propose two functions that extend the DSC, one based on absolute probability differences and one based on the Aitchison distance.  These provide a robust and accurate measure of multi-region probabilistic segmentation accuracy.
\end{abstract}

\section{Introduction}
Rigourous validation of automated and semi-automated image segmentation methods is of undeniable importance.  Aside from a few exceptions \cite{miua_2006_rss, zhang-spie, zhang2006-cvpr}, the most common validation approach in image analysis has been to compare the automated results to ``ground truth'' data, i.e. segmentations obtained by expert users or through physical or mathematical phantoms.  Thus, evaluating the accuracy of image segmentation results is typically carried out using the well-known Dice similarity coefficient (DSC) \cite{Dice1945} or other metrics such as the Hausdorff distance or the Jaccard index \cite{Jaccard1901}. Even the evaluation of image registration results is increasingly being done by evaluating the accuracy of atlas-based segmentation \cite{klein2009evaluation}, in which the DSC is commonly used.

Numerous sources of uncertainties exist in shape boundaries \cite{Udupa_prl2002}, including region heterogeneity (in medical imaging), image acquisition artifacts (e.g. blurring), and segmentation by multiple-raters. In the past decade, there has been a notable focus on encoding uncertainty in the segmentation results and not ignoring these uncertainties in subsequent analyses and decision-making \cite{Udupa_prl2002,warfield_STAPLE_2004}. In order to capture uncertainty information about the location of multiple structures in the same image, there have been several works on creating segmentation representations for multi-region probabilistic segmentations \cite{andrews2011convex}, e.g. using hyper-spherical labels \cite{Babalola_isbi2006}, the LogOdds \cite{Pohl_mia2007} of signed distance maps (SDMs) \cite{Leventon2000}, (barycentric) label-space \cite{malcolm_miccai2008,miccai2009_pmmia}, and isometric log-ratio maps \cite{changizi-miccai2010a, andrews2014isometric}. Several segmentation algorithms were also designed to use these representations to output probabilistic segmentations, both binary \cite{zhang_tmi2001,Kohli06measuringuncertainty} and multi-region \cite{Grady2006}. Further, fuzziness and uncertainty have been incorporated in other image processing and pattern recognition methods, e.g. fuzzy distance transforms \cite{Svensson-PRL07} and moments \cite{Sladoje_fuzzy2007}.  Speaking to the importance of handling probabilistic data, the visualization community has identified uncertainty visualization as a key problem in the field \cite{lundstrom_tvcg2007,Smedby_spie1999, eurovis2010, tvcg2010}. In image registration and shape matching, uncertainty calculation, visualization and utilization has also been increasingly popular \cite{miccai_miar2013, miccai_mlmi2013b, miccai2013a, miccai2012}. 

When validating a multi-region automated segmentation, be it probabilistic or crisp (non-probabilistic), a major setback is that there may be no guarantee that the segmentation will have the same number of regions as the ground truth, much less correctly corresponding labels. For example, in \cite{shi2002normalized}, Shi and Malik employ a recursive sub-optimal approach to segment multiple regions, which entails deciding if the current partition should be further sub-divided and then repartitioning if necessary. In a somewhat reverse approach, Felzenszwalb and Huttenlocher's algorithm assigns a different label to each vertex, then similar pixels are merged using a greedy decision approach \cite{felzenszwalb2004efficient}. These methods and many more \cite{coleman2005image,vincent2002watersheds} do not guarantee a particular label ordering in the resulting segmentation. Therefore, in order to properly validate an automated multi-region segmentation, label correspondences must first be determined and over- or under-segmentations handled properly.

From the previous discussion, it is evident that uncertainty-encoding segmentations that accommodate multiple-regions (with unordered labels) is an important area. However, there is a lack of published works on methods for evaluating these multi-region probabilistic segmentation results. The focus of this paper is to develop an extension of the DSC to multi-region probabilistic segmentation, along with a method for establishing label correspondences.  Specifically, we use bipartite graph matching to establish label correspondences and propose two functions that extend the DSC, one based on absolute probability differences and one based on the Aitchison distance \cite{Aitchison86}. As demonstrated by our results, we provide a robust and accurate measure for multi-region probabilistic segmentation accuracy.

\section{Method}

In Section \ref{SEC:Classical} we review the classical use of the Dice similarity coefficient (DSC) for comparing binary segmentations and discuss the challenges in extending this method to multi-region segmentation.  In Section \ref{SEC:MR} we introduce an alternate method for comparing segmentations using the DSC and then show this method extends easily to multi-region segmentation.  In Section \ref{SEC:Prob} we introduce two continuous extensions to the DSC that both allow the comparison of multi-region probabilistic segmentations and reduce to the discrete DSC when probabilities are crisp ($0$ or $1$).  Finally in Section \ref{SEC:Graphs} we propose a method for establishing label correspondences using bipartite graph matching.

\subsection{Classical Dice similarity coefficient}
\label{SEC:Classical}

The DSC measures the similarity between two sets, $X$ and $Y$ \cite{Dice1945}:
\begin{align}
	D(X,Y) = \frac{2 \left|X \cap Y \right|}{\left| X \right| + \left| Y \right|} \;. 
	\label{classical_dice}
\end{align}
where $\left| X \right|$ denotes the cardinality of the set $X$. $D(X,Y) \in [0,1]$, with $D(X,Y) = 0$ if and only if the sets are disjoint and $D(X,Y) = 1$ if and only if the sets are identical.

The DSC has been adapted to image segmentation and is a popular method for comparing binary segmentations of the same image to each other.  Often, the comparison is done between the ground truth segmentation and the results of automated or semi-automated segmentation methods.  

To measure the DSC between two segmentations, a set has to be constructed for each.  To start, one region in each segmentation is designated the foreground (as opposed to the background).  If $\Omega$ is the set of all pixels in the image, then the sets compared with the DSC are $S_1^A, S_1^{GT} \subseteq \Omega$, where $S_1^A$ is the set of pixels assigned to the foreground by the automated method and $S_1^{GT}$ is the set of pixels assigned to the foreground in the ground truth.  $D(S_1^A,S_1^{GT})$ provides a measure of how accurate an automated segmentation result is, with values closer to $1$ indicating greater accuracy.  As a simple example with $4$ pixels, $\{x_1, \cdots, x_4\}$, if $S_1^A = \{x_1,x_2,x_3\}$ and $S_1^{GT} = \{x_1,x_3\}$, then the DSC between the automated and ground truth segmentations is found using \eqref{classical_dice}: $D(S_1^A, S_1^{GT}) = \frac{2 \cdot 2}{3 + 2} = \frac{4}{5}$.

The above definition of $S_1^A$ and $S_1^{GT}$ is dependent on which region is assigned to be the foreground.  The foreground is often chosen to be the region of greatest interest, but this choice is not clear for all images, and may be dependent on the task requiring the segmentation.  Thus, when the choice of the foreground region is not clear, the DSC suffers from ambiguity as its value differs depending on this choice.  While not usually problematic for binary segmentation, when an image is segmented into multiple regions, which region to assign to the background becomes less clear, thus we would like to address this issue when extending the DSC to compare multi-region segmentations.

\subsection{Similarity coefficient for multi-region crisp segmentations} \label{SEC:MR}

Here we propose a method for using the classical DSC to compare a multi-region automated segmentation with the ground truth.  At the same time, we remove the need to specify foreground and background regions.  For now, we assume both ground truth and automated segmentations have the same number of regions. Specifically, we will assume the regions are labeled from $\mathcal{L} = \{1, \dots , L \}$, where $L$ is the number of regions, and that each region in the automated segmentation is labeled with the same number as the corresponding region in the ground truth.  This assumption will be addressed in Section \ref{SEC:Graphs}.  

\begin{figure}
	  \centering
	  	\includegraphics[width=2.5in]{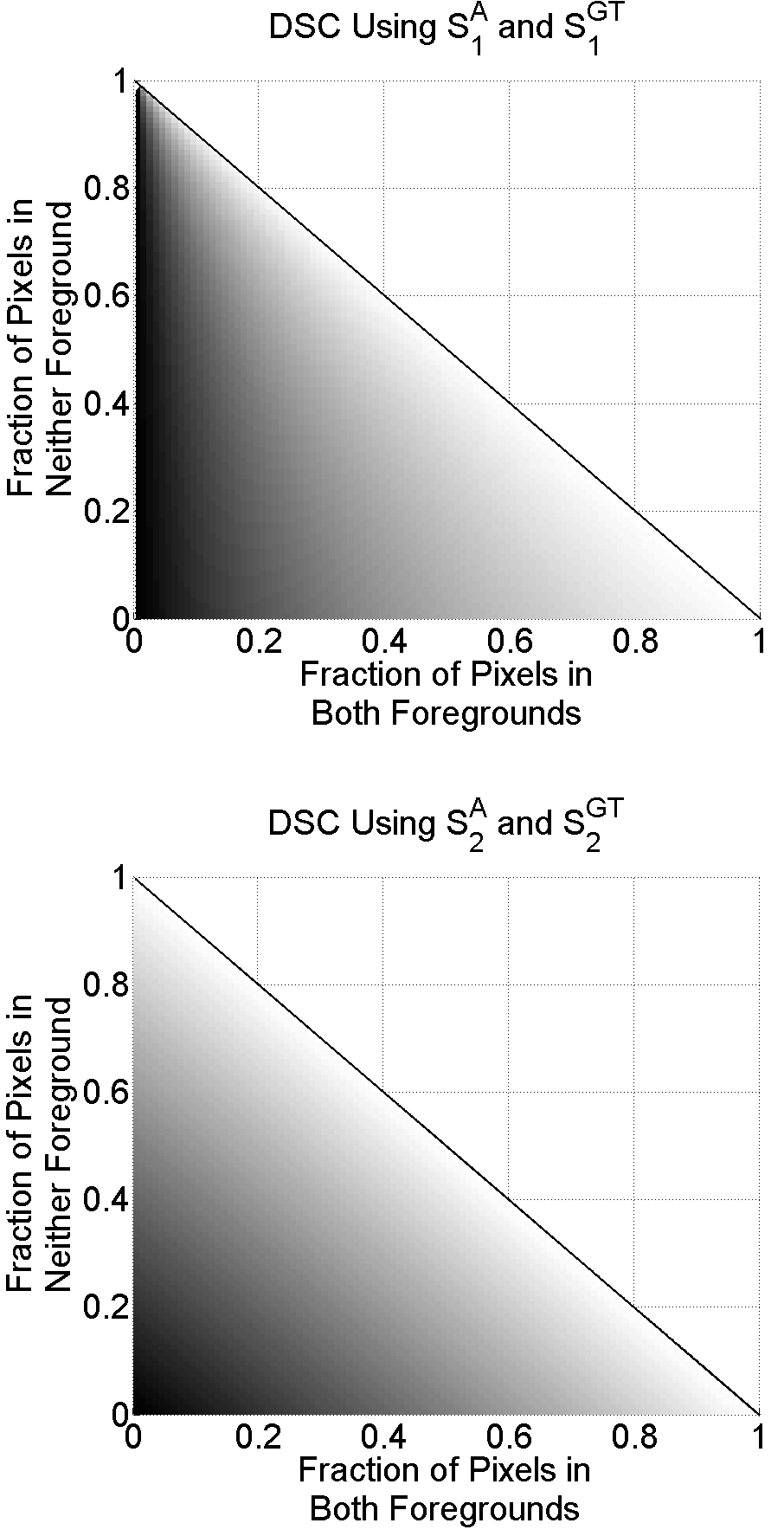} \\
	    
  \caption{\small A comparison of the DSC when using $S_1^A$ and $S_1^{GT}$ versus using our proposed $S_2^A$ and $S_2^{GT}$.  This figure assumes binary segmentations. The axes represent the fraction of pixels in the foreground and the background in both the automated and ground truth segmentations.  White corresponds to $1$ and black to $0$. Note that at the point $(0,1)$, corresponding to both segmentations being all background and thus identical, the top plot is undefined, and varies rapidly near that point.  Our approach in the bottom plot, however, assigns the value $1$ to the point $(0,1)$, correctly indicating that both segmentations are the same there.}
	  \label{fig:Dice_comp}
\end{figure}

Again, we must construct a set for each segmentation.  Here, we propose the sets $S_2^A, S_2^{GT} \subset (\Omega \times \mathcal{L})$, that is, sets of ordered pairs consisting of a pixel and an integer from $1$ to $L$.  The set $S_2^A$ will have, for each pixel, an element containing the label of the region that that pixel is assigned to in the automated segmentation; $S_2^{GT}$ will be defined similarily for the ground truth.  Thus we have $\left| S_2^A \right| = \left| S_2^{GT} \right| = \left| \Omega \right|$.  By comparing these sets using the DSC we get a value from $0$ to $1$ indicating what fraction of pixels share a label in both segmentations.  As a simple example with $4$ pixels, $\{x_1, \cdots, x_4\}$ and $2$ regions, labeled from $\mathcal{L} = \{1,2\}$, if $S_2^A = \left\{\{x_1,1\},\{x_2,1\},\{x_3,1\},\{x_4,2\}\right\}$ and $S_2^{GT} = \left\{\{x_1,1\},\{x_2,2\},\{x_3,1\},\{x_4,2\}\right\}$, then the DSC between the automated and ground truth segmentations is found using \eqref{classical_dice}: $D(S_2^A, S_2^{GT}) = \frac{2 \cdot 3}{4 + 4} = \frac{3}{4}$.

For binary segmentation ($L = 2$) this definition of $S_2^A$ and $S_2^{GT}$ makes the DSC take the value $1$ only when the segmentations are identical, and $0$ when none of the pixels have the correct (ground truth) label. Furthermore, both this method and the method introduced in Section \ref{SEC:Classical} for calculating the DSC increase as the number of pixels assigned to the same region (foreground or background) in both segmentations increases (Figure \ref{fig:Dice_comp}).  Thus, $S_2^A$ and $S_2^{GT}$ give a comparison metric similar to $S_1^A$ and $S_1^{GT}$, yet do not suffer from the aforementioned ambiguity and extend naturally to multi-region segmentations.

\subsection{Similarity coefficient for multi-region probabilistic segmentations} \label{SEC:Prob}

When comparing discrete, non-probabilistic segmentations, similarity is justifiably measured in a discrete way: whether or not a pixel is assigned to the same region in both segmentations.  Thus constructing sets and applying the classical DSC \eqref{classical_dice}, as in Sections \ref{SEC:Classical} and \ref{SEC:MR}, accurately captures the similarity between segmentations.  However, when considering continuous, probabilistic segmentations, such discrete comparisons are no longer applicable.  

For example, if the ground truth segmentation assigns pixel $x$ probability $0.9$ of being in region $r$, then an automated segmentation that assigns pixel $x$ probability $0.7$ of being in region $r$ should be considered more similar to the ground truth than if it had assigned probability $0.6$, but less similar than if it had assigned probability $0.8$.

To accurately capture the similarity between multi-region probabilistic segmentations, we need to extend the DSC to a continuous function.  We will again assume that the automated and ground truth segmentations have the same number of regions, labeled from $\mathcal{L} = \{1, \dots , L \}$, with corresponding regions labeled the same (violating this assumption is addressed in Section \ref{SEC:Graphs}).  We define the simplex of probability vectors of length $L$:
\begin{align}
	\mathbb{S}^{L}= &\left\{ [p_1, p_2, . . . , p_L] \in \mathbb{R}^L \phantom{\sum_{i=1}^L p_i}\right.  \notag\\ 
	%\mathbb{S}^{k}=\left\lbrace &[p_1, p_2, . . . , p_D] \in \mathbb{R}^D \right. \; \vert  \\ 
	   & \left. p_i \geq 0 , i=1,2,\dots, L  \; ; \sum_{i=1}^L p_i =1 \right\} \;.
\end{align}

We let $p^A$ and $p^{GT}$ represent the multi-region probabilistic automated and ground truth segmentations, respectively. $p^A$ and $p^{GT}$ assign to each pixel $x$ regional (or label) probabilities $p^A(x), p^{GT}(x) \in \mathbb{S}^{L}$.  Our first step will be to define a pixel to pixel similarity function $f: \mathbb{S}^{L} \times  \mathbb{S}^{L} \to [0,1]$, mapping two vectors of $L$ regional probabilities to the interval $[0,1]$ such that larger values correspond to more similar regional probabilities.  Furthermore, $f = 1$ should imply the regional probabilties are identical and $f=0$ should imply that every region is assigned probability $0$ by at least one of the segmentations.

Once we have defined $f$, we can define a continuous extension to the DSC, $D^{cts}$, that compares two multi-region probabilistic segmentations:
\begin{align}
	D^{cts}(p^A, p^{GT}) = \frac{1}{\left| \Omega \right|} \sum_{x \in \Omega} f \left(p^A(x), p^{GT}(x) \right) \;. \label{cts-dice}
\end{align}
Since $f \in [0,1]$, dividing the summation by $\left| \Omega \right|$ ensures $D^{cts} \in [0,1]$. $D^{cts}=1$ if and only if $f=1$ for all pixels and $D^{cts}=0$ if and only if $f=0$ for all pixels.  Note that, given our requirements for $f$, $D^{cts}(p^A, p^{GT})$ reduces to $D(S_2^A, S_2^{GT})$ from \eqref{classical_dice} in the case that all probabilities are either $0$ or $1$.

\begin{figure}
	  \centering
	  	\includegraphics[width=2.5in]{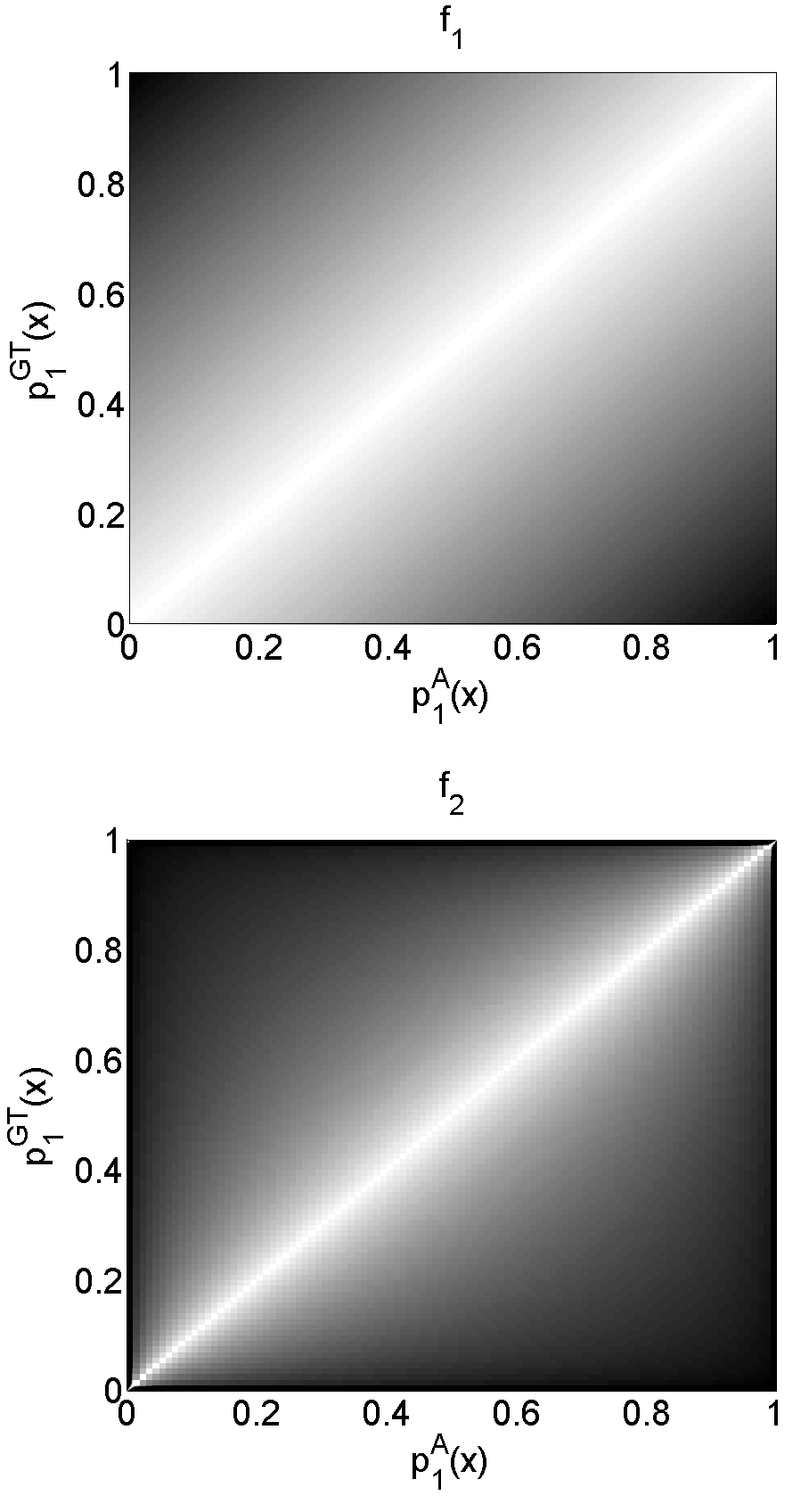} \\
	    
  \caption{\small A comparison of $f_1$ and $f_2$ for varying regional probabilities at a given pixel and $L=2$ regions.  The $x$ and $y$ axes represent the probability of the pixel belonging to the first region in the automated and ground truth segmentations, respectively. Note how $f_2$ drops off more quickly near the diagonal and drops to $0$ around the boundaries, when one of the segmentations is almost certain (see discussion in the text). White corresponds to $1$ and black to $0$.}
	  \label{fig:f_comp}
\end{figure}

We propose two versions of $f$, and we will discuss later some situations when one is more applicable than the other.  We define the first version by looking at the absolute value of the difference between regional probabilities in the two segmentations:
\begin{align}
	f_1(p^A(x), p^{GT}(x)) = 1 - \frac{1}{2} \sum_{i=1}^{L} \left| p^A_i(x) - p_i^{GT}(x) \right| \;, \label{f1}
\end{align}
where $x \in \Omega$. $f_1 = 1$ if and only if $p^A(x) = p^{GT}(x)$. $f_1 = 0$ if and only if every region is assigned probability $0$ by at least one of the segmentations.

The case of $f_1 = 0$ can be seen by first noting that the $\left| p^A_i(x) - p_i^{GT}(x) \right| \leq \max(p^A_i(x),p_i^{GT}(x))$, with equality holding only when one of the probabilities is $0$.  Thus, if neither of the probabilities are $0$ for some region,
\begin{align}
	\sum_{i=1}^L p^A_i(x) + \sum_{i=1}^L p_i^{GT}(x) = 2 &=  \sum_{i=1}^L p^A_i(x) + p_i^{GT}(x) \\
	&\geq \sum_{i=1}^L \max(p^A_i(x), p_i^{GT}(x)) \\
	&> \sum_{i=1}^{L} \left| p^A_i(x) - p_i^{GT}(x) \right| \\
	&= -2(f_1(x)-1)\;,
\end{align}
i.e. $f_1 > \frac{2}{(-2)}+1 $, which implies $f_1 > 0$.  When all regions are assigned probability $0$ by at least one of the segmentations, $p^A_i(x) + p_i^{GT}(x) = \left| p^A_i(x) - p_i^{GT}(x) \right|$ and thus $f_1 = 0$.

Substituting $f_1$ for $f$ in \eqref{cts-dice} gives a function $D_1^{cts}$ that extends the DSC to multi-region probabilistic segmentations and reduces to the discrete DSC from Section \ref{SEC:MR} when all probabilities are either $0$ and $1$.

Although $f_1$ achieves our goal, the space $\mathbb{S}^L$ is a Hilbert space and as such has an inner product defined on it.  This inner product induces a distance function $d_a: \left( \mathbb{S}^L \right)^2 \to \mathbb{R}^+$, known as the Aitchison distance \cite{Aitchison86}:
\begin{align}
	d_a(p, q) &= \left[\sum_{\ell=1}^L\left(\log \frac{p_\ell}{\mu_g(p)}- \log \frac{q_\ell}{\mu_g(q)}\right)^2 \right]^{\frac{1}{2}} \;, \label{aitch-dist}
\end{align}
where $p,q \in \mathbb{S}^L$ with components $p_\ell$ and $q_\ell$, and $\mu_g$ is the geometric mean.  We can make use of the Aitchison distance to create another version of $f$, denoted $f_2$, that utilizes this more natural way to compare probability vectors.  Now, since $d_a$ is a distance function, it is $0$ when the probability vectors being compared are identical, and approaches $\infty$ as the probability vectors become maximally different.  Thus, we define
\begin{align}
	f_2(p^A(x), p^{GT}(x)) = 	\frac{1}{1 + d_a(p^A(x), p^{GT}(x))} \;, \label{f2}
\end{align}
where $x \in \Omega$.  Note that $f_2 = 1$ when $d_a = 0$ and $f_2 \to 0$ as $d_a \to \infty$, and also that 
\begin{align}
	\lim_{\alpha \to 1} d_a([\alpha,1-\alpha,0,\dots], [1-\alpha,\alpha,0,\dots]) &= \infty \\
	\Rightarrow \quad f_2([1,0,0,\dots], [0,1,0,\dots]) &= 0 \;,
\end{align} 
for $[\alpha,1-\alpha,0,\dots] \in \mathbb{S}^L$.  Substituting $f_2$ for $f$ in \eqref{cts-dice} gives a function $D_2^{cts}$ that, as $D_1^{cts}$, extends the DSC from Section \ref{SEC:MR} and reduces to it for discrete segmentations.  

\comment{
\begin{figure}
	  \centering
	  	\subfloat[Seg. $1$]{\label{fig:eta-en1}\includegraphics[width=1.5in]{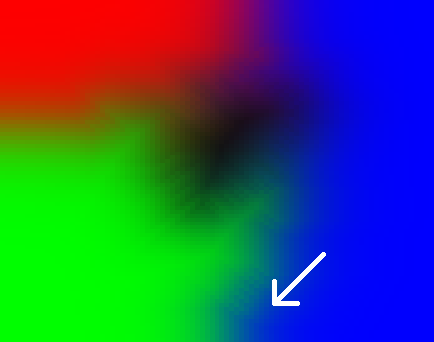}} \;
	    \subfloat[Seg. $2$]{\label{fig:eta-en1}\includegraphics[width=1.5in]{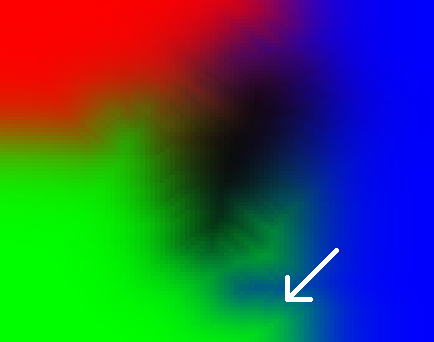}}\\
	    \subfloat[$f_1$]{\label{fig:eta-en2}\includegraphics[width=1.5in]{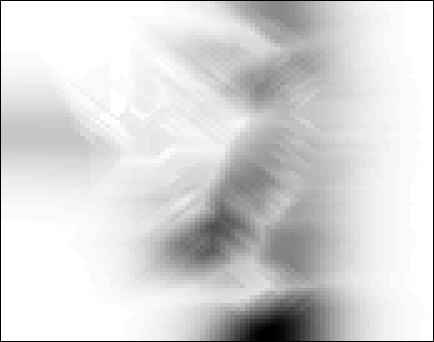}} \;
	    \subfloat[$f_2$]{\label{fig:eta-en2}\includegraphics[width=1.5in]{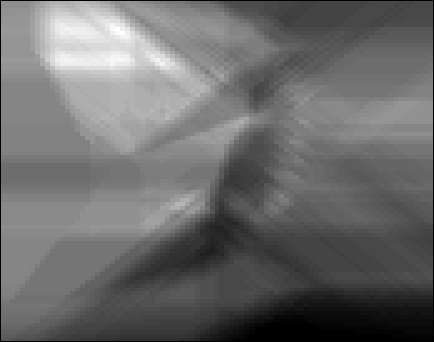}} \\
  \caption{\small (Colour Figure) Two probabilistic segmentations of the same image and their values of $f_1$ and $f_2$ at each pixel.  In the figures showing the segmentations, the RGB values correspond to the probabilities of three of the regions.  In the figures for $f_1$ and $f_2$, white corresponds to $1$ and black to $0$. We note the greatest difference between the segmentations is in the bottom middle (white arrow), where one segmentation favors the blue region and the other the green.  We also note that while $f_1$ takes lower values mostly around region boundaries where the segmentations differ, $f_2$ is \ednote{more sensitive *** yes, can you use less different examples?} to the smaller differences across the segmentations. \ednote{with Figure \ref{fig:comp_ex}, is this figure still needed?}}
	  \label{fig:f_ex}
\end{figure}}
\captionsetup[subfloat]{labelformat=empty}

\begin{figure}
	  \centering
%	  	\subfloat[Ground Truth]{\label{fig:x}\includegraphics[width=1.0in]{Images/gt_seg.png}} \;
%	    \subfloat{\label{fig:x}\includegraphics[width=1.0in]{Images/base.png}}\;
%	    \subfloat{\label{fig:x}\includegraphics[width=1.0in]{Images/base.png}} \\
%	    \subfloat[Segmentation $1$]{\label{fig:x}\includegraphics[width=1.0in]{Images/a_seg1.png}} \;
%	    \subfloat[$f_1$: DSC $= 0.7941$]{\label{fig:x}\includegraphics[width=1.0in]{Images/f11.png}}\;
%	    \subfloat[$f_2$: DSC $= 0.7220$]{\label{fig:x}\includegraphics[width=1.0in]{Images/f21.png}} \\
			\subfloat[Ground Truth]{\label{fig:x}\includegraphics[width=0.7in]{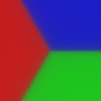}} \;
	    \subfloat[Segmentation $1$]{\label{fig:x}\includegraphics[width=0.7in]{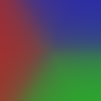}} \;
	    \subfloat[$f_1$: DSC $= 0.5807$]{\label{fig:x}\includegraphics[width=0.7in]{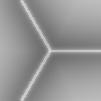}}\;
	    \subfloat[$f_2$: DSC $= 0.5707$]{\label{fig:x}\includegraphics[width=0.7in]{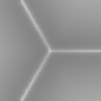}} \\
	    \subfloat{\label{fig:x}\includegraphics[width=0.7in]{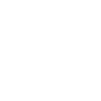}}\;
	    \subfloat[Segmentation $2$]{\label{fig:x}\includegraphics[width=0.7in]{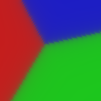}} \;
	    \subfloat[$f_1$: DSC $= 0.8339$]{\label{fig:x}\includegraphics[width=0.7in]{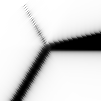}}\;
	    \subfloat[$f_2$: DSC $= 0.8429$]{\label{fig:x}\includegraphics[width=0.7in]{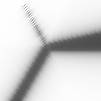}} \\
  \caption{\small (Colour Figure) An example showing $f_1$ and $f_2$ at each pixel when comparing the top left segmentation (Ground Truth) to each of the other segmentations ($1$ and $2$). In the segmentations, the RGB values correspond to the probabilities of the three regions.  In the figures for $f_1$ and $f_2$, white corresponds to $1$ and black to $0$.}
	  \label{fig:synth_fs}
\end{figure}

Thus we have two functions, $D_1^{cts}$ and $D_2^{cts}$ that extend the DSC to multi-region probabilistic segmentations.  To compare $D_1^{cts}$ and $D_2^{cts}$, we compare $f_1$ and $f_2$ at a pixel $x$.  We will hold the regional probabilities of the ground truth, $p^{GT}(x)$, fixed and consider changing the regional probabilities of the automated segmentation, $p^A(x)$.  The values of $f_1$ and $f_2$ at a single pixel with varying probabilties is seen in Figure \ref{fig:f_comp}.  As $p^A(x)$ changes, $f_1(x)$ varies linearly with the individual probabilities, while $f_2(x)$ varies more rapidly when $p^A(x)$ is close to $p^{GT}(x)$.  However, due to the nature of the Aitchison distance, $f_2(p^A(x), p^{GT}(x)) = 0$ if either $p^A(x)$ or $p^{GT}(x)$ is certain (i.e. contains a $0$ or a $1$). This behaviour is reasonable when we consider that having \emph{absolute certainty} in a pixel's label (even from manual segmentation) is arguably unachievable in reality (see Cromwell's rule \cite{jackman2009bayesian}).

Thus in applications when the automated segmentations are likely to be close to the ground truth, $D_2^{cts}$ will be more sensitive to small differences, but in applications that are likely to have many completely certain pixels, $D_1^{cts}$ will better capture segmentation differences.   We see a comparison of $f_1$ and $f_2$ applied to two probabilistic segmentations in Figure \ref{fig:synth_fs}. 

\begin{figure}[t]
	  \centering
	  	\includegraphics[width=0.45\textwidth]{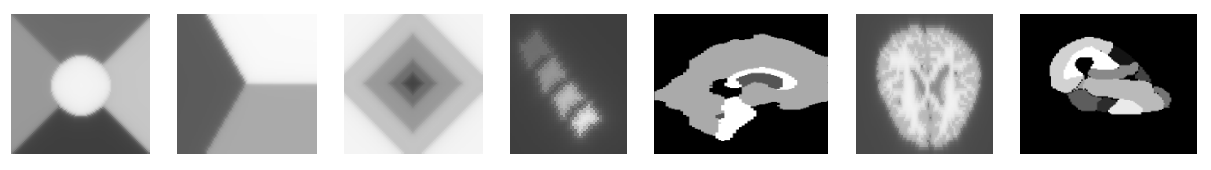} \\
  \caption{\small All $7$ images deformed and used in the user study summarized in Figure \ref{fig:survey}.}
	  \label{fig:all_gts}
\end{figure}

\begin{figure*}
	  \centering
	  \includegraphics[width=\textwidth]{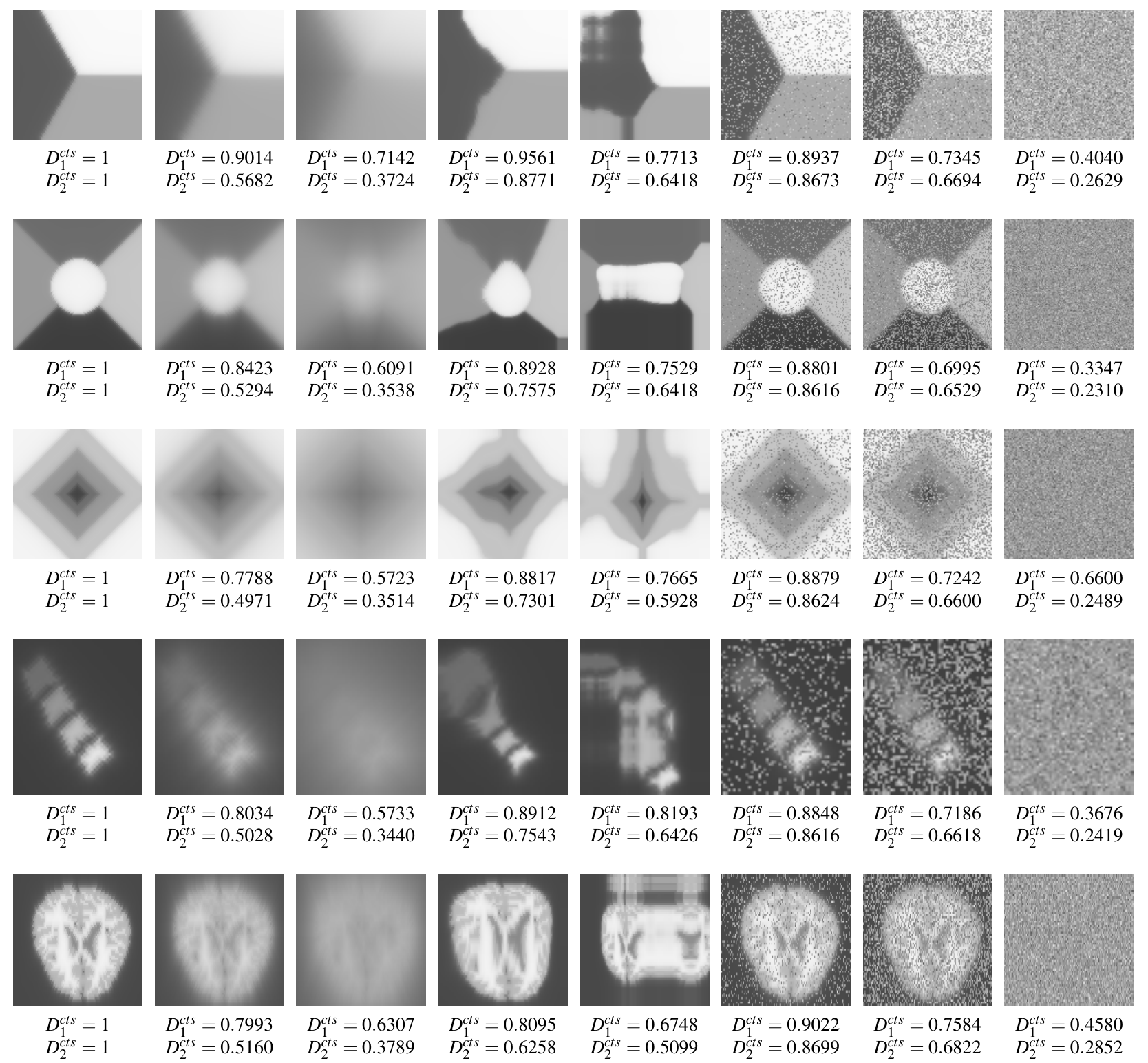} \\
	  		    
  \caption{\small A comparison of various ground truth segmentations (left) and example segmentations.  The multi-region probabilistic ground truth segmentations for five images, both real and synthetic, are seen on the left.  These are compared to segmentations generated from their ground truths by blurring the probabilities, spatial deformations, and the addition of noise.  The corresponding DSCs, both $D_1^{cts}$ and $D_2^{cts}$, are reported for each segmentation. As expected, both DSCs drop as the segmentation deviates further from the ground truth.}
	  \label{fig:comp_ex}
\end{figure*}

\begin{figure}[t]
	  \centering
	  	\includegraphics[width=0.45\textwidth]{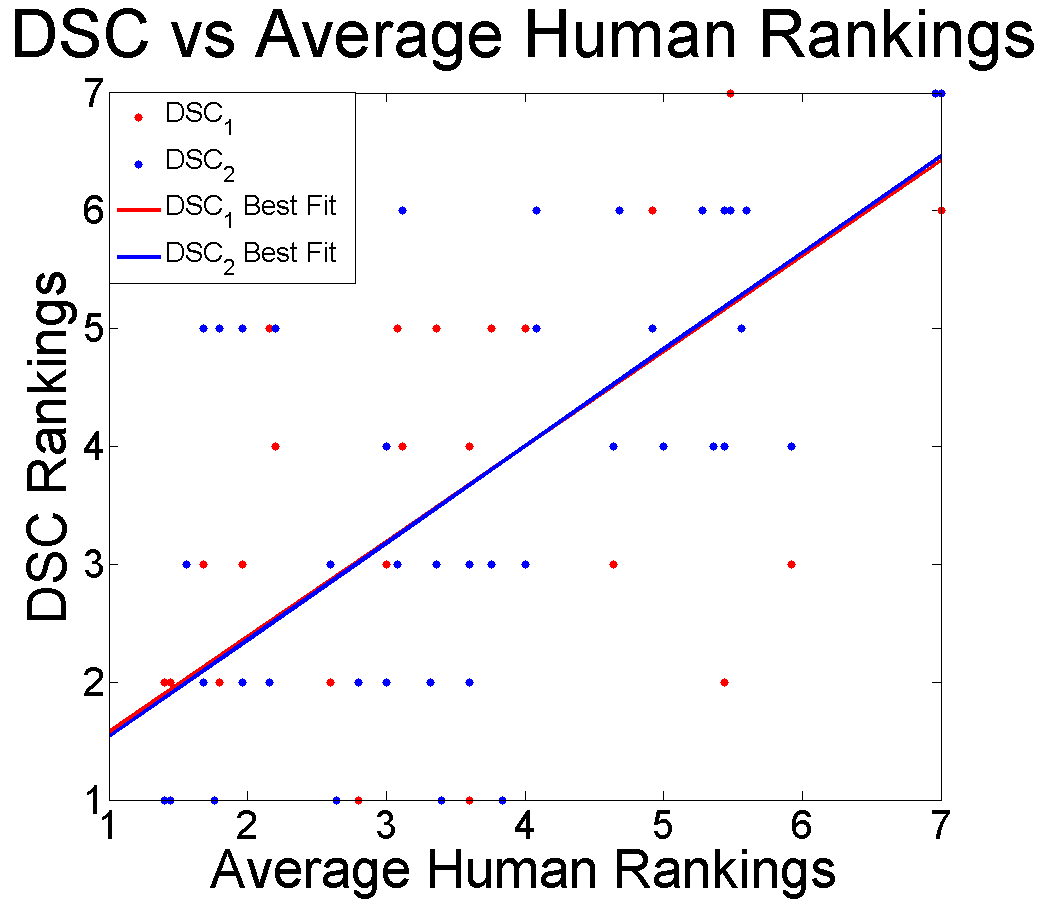} \\
  \caption{\small A comparison of average human rankings of the $49$ incorrect segmentations (see Figure \ref{fig:comp_ex}) and the DSC ranks.  We see a strong correlation between the human rankings and the DSC rankings, with Pearson's correlation coefficient of $0.72$.}
	  \label{fig:survey}
\end{figure}

To analyze the performance of our method, we carry out the following user study: Twenty five people (unaware of the purpose of the study) were given a random ordering of $7$ incorrect segmentations of the same image and the corresponding GT segmentation, and asked to rank the $7$ incorrect segmentations from most to least similar to the GT, where $1$ indicates most similar and $7$ least similar.  The incorrect segmentations were generated using blurring, spatial deformations, and noise on the images in Figure \ref{fig:all_gts}.  $5$ of the sets of segmentations and their values of $D_1^{cts}$ and $D_2^{cts}$ with respect to the GT are seen in Figure \ref{fig:comp_ex}. In Figure \ref{fig:survey}, we see the results of the survey, with each point corresponding to an incorrect segmentation, the $x$-axis corresponding to the average human ranking, and the $y$-axis corresponding to the ranking given by the proposed DSC.  The line of best fit through the data has a slope of $0.82$, indicating a strong correlation.  Furthermore, Pearson's correlation coefficient between the DSC rankings and the average human rankings was $0.72$.

\begin{figure}
	  \centering
	  	\includegraphics[width=2.5in]{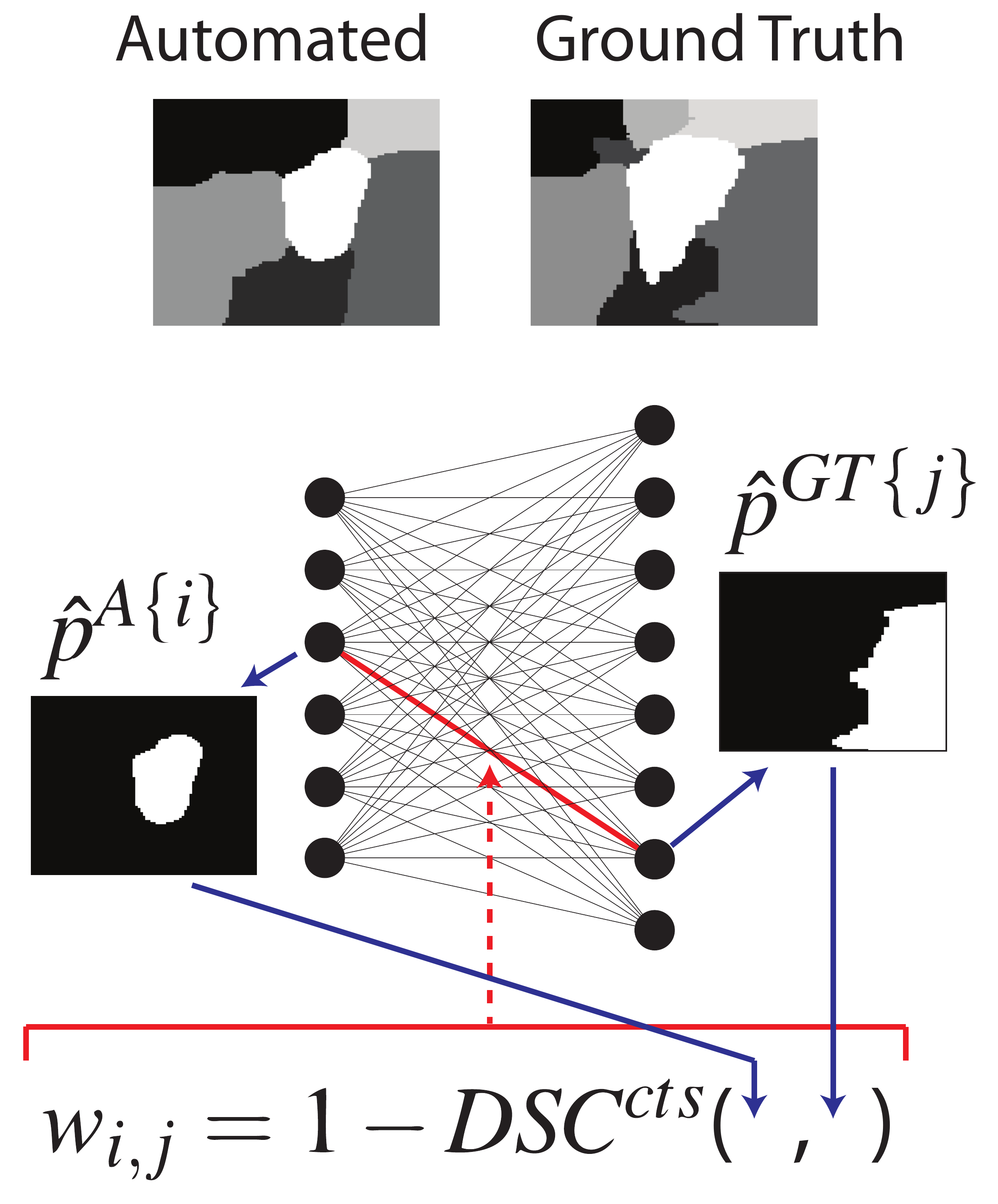} \\
  \caption{\small (Colour Figure) An example showing how the bipartite graph is constructed and edge weights calculated. The edge between two nodes is highlighted in red and the auxiliary binary segmentations for those two nodes is shown.  These auxillary segmentations are compared to find a weight for the edge.}
	  \label{fig:graph_schem}
\end{figure}

\captionsetup[subfloat]{labelformat=empty}

\subsection{Finding label correspondences via graph matching} \label{SEC:Graphs}

There is no guarantee that an automated multi-region segmentation method will label its regions in a way that correspond to the labels of the ground truth's regions.  This correspondence, however, is required to compare the segmentations using the DSC and extensions described in Sections \ref{SEC:MR} and \ref{SEC:Prob}.  In fact, if $\mathcal{L}_1$ is the set of region labels from the automated segmentation and $\mathcal{L}_2$ is the set of region labels from the ground truth, there is no guarantee that even $\left| \mathcal{L}_1 \right| = \left| \mathcal{L}_2 \right|$.  We require a way of establishing a correspondence between the regions in the two segmentations.  

In this section we assume the segmentations are probabilistic, as these are a superset of discrete segmentations and Section \ref{SEC:Prob} extends the discrete methods of Section \ref{SEC:MR} to continuous methods.

We establish label correspondences by constructing a weighted complete bipartite graph, with $L_1 = \left| \mathcal{L}_1 \right|$ vertices on the left representing the $L_1$ regions in the automated segmentation and $L_2 = \left| \mathcal{L}_2 \right|$ vertices on the right representing the $L_2$ regions in the ground truth.  For each pair of regions $(i,j) \in (\mathcal{L}_1 \times \mathcal{L}_2)$ we will assign a weight $w_{i,j}$ to the edge between their corresponding vertices.  We want $w_{i,j}$ to be smaller when regions $i$ and $j$ match well, and larger when they match poorly.  Thus, to calculate $w_{i,j}$, we create an auxiliary binary segmentation, $\hat{p}^{A\{i\}}(x)$, from the automated  segmentation $p^A(x)$, by treating region $i$ as the foreground and all other regions in the automated segmentation as the background. Specifically, the probabilities of $\hat{p}^{A\{i\}}$ at a pixel $x$ are given by  
\begin{align}
	\hat{p}^{A\{i\}}_1(x) &= p^A_i(x) \\
	\hat{p}^{A\{i\}}_2(x) &= \sum_{\substack{\ell \in \mathcal{L}_1 \\ \ell \neq i}} p^A_\ell(x) \;.
\end{align}
We create a similar auxillary segmentation $\hat{p}^{GT\{j\}}$ using region $j$ in the ground truth segmentation.  

We then calculate the DSC between the auxillary segmentations using\footnote{Using either $D_1^{cts}$ or $D_2^{cts}$ depending on the application.} $D^{cts}$ from \eqref{cts-dice}.  Doing so will give a similarity coefficient between regions $i$ and $j$ that is independent of the other regions. The edge in the bipartite graph between the vertices corresponding to regions $i$ and $j$ will then be assigned the weight $w_{i,j} = 1-D^{cts}$. Figure \ref{fig:graph_schem} illustrates an example of how an edge weight is calculated.

Once the graph is constructed, we apply the Hungarian (Kuhn-Munkres) bipartite graph matching algorithm to find a minimal matching \cite{kuhn1955}. We take this matching as the correspondence between the regions of the two segmentations. We use this correspondence to relabel the automated segmentation.  

\begin{figure}[h]
	  \centering
	  	\subfloat[Region Correspondences]{\label{fig:eta-en1}\includegraphics[width=2.5in]{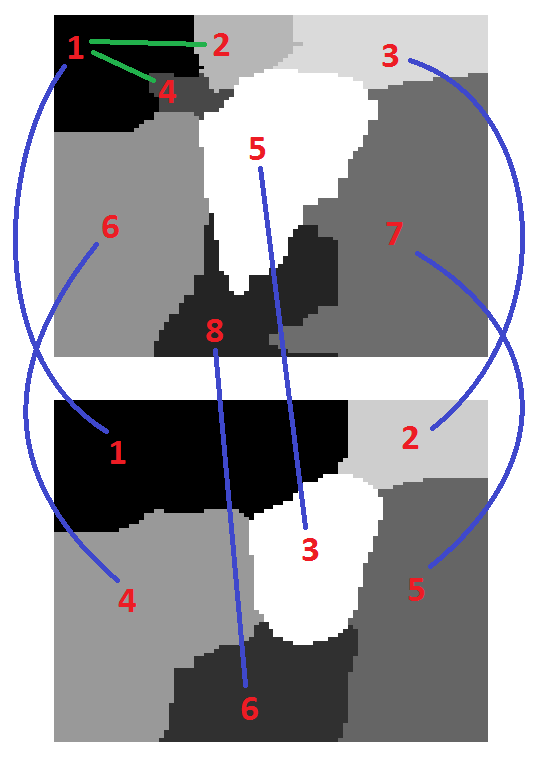}} \\
  \caption{\small (Colour Figure) Two segmentations of the same image and the correspondences established using our approach.  Blue lines indicate regions matched between the segmentations using the bipartite graph and green lines indicate which matched region unmatched regions were added to. }
	  \label{fig:reg_corr}
\end{figure}

When $\left| \mathcal{L}_1 \right| > \left| \mathcal{L}_2 \right|$, it may be the case that some regions were oversegmented, and multiple regions from the automated segmentation all correspond to the same region in the ground truth. Therefore, we may wish to merge regions together to obtain a more optimistic estimate of the DSC. If this is the case, after the initial matching let $\mathcal{M}$ and $\mathcal{U}$ be the regions of the automated segmentation that are matched or unmatched to regions in the ground truth, respectively.  We want to add each of the regions from $\mathcal{U}$ to one of the regions from $\mathcal{M}$.  Given an unmatched region from $\mathcal{U}$ (denoted $i_u$), for each pair of matched regions from $(\mathcal{M} \times \mathcal{L}_2)$ (denoted $i_m$ and $j_m$) we will create a new auxillary binary segmentation $\hat{p}^{A\{i_u,i_m\}}$ by treating the combination of regions $i_u$ and $i_m$ as the foreground and all other regions in the automated segmentation as the background.  Specifically, the probabilities of $\hat{p}^{A\{i_u,i_m\}}$ at a pixel $x$ are given by 
\begin{align}
	\hat{p}^{A\{i_u,i_m\}}_1(x) &= p^A_{i_u}(x) + p^A_{i_m}(x) \\
	\hat{p}^{A\{i_u,i_m\}}_2(x) &= \sum_{\substack{\ell \in \mathcal{L}_1 \\ \ell \neq i_u, i_m}} p^A_\ell(x) \;.
\end{align}

\noindent We will then decide if adding $i_u$ to $i_m$ improves the matching with $j_m$ by taking the difference  
\begin{align}
	\delta_{i_u}(i_m) = D^{cts} \left( \hat{p}^{A\{i_u,i_m\}}, \hat{p}^{GT\{j_m\}} \right) - \notag \\ 
	D^{cts} \left( \hat{p}^{A\{i_m\}}, \hat{p}^{GT\{j_m\}} \right) \; .
\end{align}

\noindent Once we have calculated $\delta_{i_u}$ for each matched region we will permanently add $i_u$ to the region given by
\begin{align}
	\argmax_{i_m} \delta_{i_u}(i_m) \;,
\end{align}
i.e. the region whose matching will be improved the most by the addition of $i_u$. We then update $\mathcal{M} = \mathcal{M} \cup i_m$, and $\mathcal{U} = \mathcal{U} \backslash i_m$ and repeat until $\mathcal{U}=\emptyset$. A similar secondary matching phase may be used if $\left| \mathcal{L}_1 \right| < \left| \mathcal{L}_2 \right|$ and the automated segmentation is thought to have under-segmented some regions. 

Figure \ref{fig:reg_corr} shows a result of our region matching technique.  All $6$ regions from the bottom segmentation are matched to the region from the top segmentation with which they have the highest DSC. For example, the DSC between region $4$ in the bottom segmentation and region $6$ in the top segmentation is $0.9370$, whereas region $4$ from the bottom has DSC less than $0.75$ with all other regions from the top segmentation.  Regions $2$ and $4$ in the top segmentation were not matched using the bipartite graph, and it was found that when they were both added to region $1$ in the top segmentation they improved the DSC with region $1$ in the bottom segmentation from $0.8757$ to $0.9329$.

Combining the techniques introduced in this section with the extended DSC function \eqref{cts-dice} enables the comparison of any two multi-region probabilistic segmentations, even if they have different numbers of regions or if one is crisp and the other probabilistic. 

\section{Conclusions}
Validation is crucial for automated and semi-automated segmentation methods, and this can be achieved by comparing the resulting segmentations with a ground truth segmentation.  Such comparisons are often done using the DSC, but this method only applies to crisp binary segmentations.  We have motivated the importance of multi-region probabilistic segmentations, and thus the importance of extending the DSC to compare such segmentations.  We have achieved this goal by proposing two different extensions with different qualities that allow the comparison of probabilistic or crisp segmentations with any number of regions in one framework.  We have shown how to establish label correspondences between segmentations, even when they have different numbers of regions, so that the DSC can be applied.  This work greatly extends the usability of the DSC and provides a seamless comparison metric across a wide variety of segmentations (e.g. crisp and probabilistic; binary and multi-region;  and differing number of regions, with and without ordered labels).

\bibliographystyle{plain}
\bibliography{multi_region_probabilistic_dice}
\end{document}